\documentstyle[conference]{article}
\def\false{|\!\!\not\models}
\begin{document}
\twocolumn
\title{Combining Expression and Content in Domains for Dialog Managers}
\author{Bernd Ludwig\and G\"unther G\"orz\and Heinrich
  Niemann\\{\tt bdludwig@forwiss.uni-erlangen.de}\\{\tt
\{goerz,niemann\}@informatik.uni-erlangen.de}\\Bavarian Research
Center for Knowledge Based Systems (FORWISS)\\Am Weichselgarten 7,
D-91058 Erlangen, Germany.}
\maketitle
\begin{abstract}
We present work in progress on abstracting dialog
managers from their domain in order to implement a dialog manager
development tool which takes (among other data) a domain description
as input and delivers a new dialog manager for the described domain as
output. Thereby we will focus on two topics; firstly, the
construction of domain descriptions with description logics and
secondly, the interpretation of utterances in a given domain.
\end{abstract}
\section{Introduction}

Current research on dialog management is guided by two different
ideas: firstly, to describe the discourse structure that the dialog manager
is able to handle by a finite automaton using possible utterances as
transitions (e.g.\ \cite{baekgaard}), and secondly, to view the
detection of discourse structure as a parameter estimation problem and
to use statistical models for the description of discourse structure
(e.g.\ \cite{moeller}, \cite{shriberg}).

Some serious problems remain with each of these approaches: first of
all, they do not integrate a user model. The finite state method
imposes hard restrictions on ``free dialog'', as far as vocabulary,
syntax and order of utterances are concerned. The statistical
approach, on the other hand, suffers from the sparse data problem and
does not give theoretical insight into discourse theory.

But as researchers point out (see e.g.\ \cite{novick}, \cite{agarwal}),
it is important to explore the structure of discourse and the
interactions of dialog and user model in order to obtain robust dialog
systems. Studying the effect of natural language expressions on the
state of discourse this paper tries to do a step in this direction by
discussing the use of description logics for defining dialog
domains formally and by describing how inference is performed on the
basis of such a framework.

Following Hjelmslev \cite{Hjelmslev1943}, Eco \cite{Eco1993}
defines two essential ingredients for any semiotic system (and, in
particular, any natural language): 
\begin{itemize}
\setlength{\itemsep}{0pt}
\setlength{\parsep}{0pt}
\item Expressiveness: what is the vocabulary, phonetics, and syntax of the
language considered?
\item Content: what know\-ledge can be expressed by a given system? How
(i.e. by which expressions) is this knowledge organised in the
semiotic system?
\end{itemize}

Obviously, the key problem is how to connect content and
expressions in order to capture the meaning of expressions. Russell
\cite{Russell1940} proposes to divide the vocabulary into two classes:
\begin{itemize}
\setlength{\itemsep}{0pt}
\setlength{\parsep}{0pt}
\item object words: define basic notions (in a train information
  scenario e.g.\ {\sl train}, {\sl time}, or {\sl station})
\item dictionary words: can be defined using other words with already
defined meaning (e.g.\ {\sl departure time} or {\sl arrival station})
\end{itemize}
 
Trivially, there is an obvious analogy between object words and
primitive concepts in description logics and dictionary words and
derived concepts. We exploit this analogy to describe the application
domain for a dialog manager by means of a terminology in DL. Doing
this, we can define the meaning of basic and derived notions that are
relevant in the domain under consideration.

\section{$\lambda$-DRT and DL}
\label{drtanddl}
A basic concept of any dialog system is a theory of discourse. We
use Discourse Representation Theory (DRT) as introduced by
\cite{reyle} to describe the discourse generated during a
dialog. DRT has simple semantics: A discourse representation structure
(DRS) $K$ with
$$
K=\left[\begin{array}{l}d_1\,d_2\,...\\\hline
\mbox{cond}_1(d_1,d_2,...)\\ 
\mbox{cond}_2(d_1,d_2,...)\\
...
\end{array}\right]
$$
can be mapped into the logic language capturing the syntax of
$\mbox{cond}_i(d_1,d_2,...)$ by:
$$
\exists d_1,d_2,...:\mbox{cond}_1(d_1,d_2,...)\wedge
\mbox{cond}_2(d_1,d_2,...)
$$
This mapping shows that a DRS essentially defines a set of assertions
and thereby expresses extensional knowledge declared throughout the
discourse. 

On the other hand, a terminology of a
certain domain can describe intensional knowledge used while
constructing DRS out of utterances: the linguistic parser has to
verify constraints imposed by subcategorization. E.g.\ a train usually
departs from a station, i.e., a city name is a valid object of {\sc
Depart} if there exists a railway station. To evaluate this constraint
the parser makes use of DL reasoning services while constructing the
semantics of the current utterance.

In this sense the incorporation of DL into our approach to dialog theory is
crucial to overcome some of the limitations\footnote{First order logic
lacks a mechanism for constructing well-defined terms from
utterances. Its model theory is intensionally and
ontologically inadequate for natural language expressions.} of first
order logic for describing natural language semantics.

\section{Partial Information in Dialogs}
\label{fil}
A key issue for dialog management is the handling of partial
information provoked by ``incomplete'' utterances. By this notion we
mean the fact that in the {\it shared knowledge} of the participants
there is not enough information available to, e.g., answer a question
posed by the speaker. For the sake of illustration, let us consider the query
\vskip.33\baselineskip
\centerline{\it When does a train leave to Rome?}
\vskip.33\baselineskip
The information given in this query is (only) partial as it is
incomplete or insufficient for the hearer to answer it. At least the
station of departure is unknown and has to be asked for by the hearer
if he wants to give some reasonable answer. Notably, the open world
assumption of DL does not suffice to handle, because it does not allow
to make an assumption about a formula (representing a query as above),
not even by default.
 
We use First Order Partial Information Ionic Logic (FIL), as
introduced by Abdallah \cite{abdallah}, to handle situations of partial
knowledge. FIL is a first order language whose interpretation function
is partial, i.e.\ a formula can have a undefined truth value. Of
course, interpretation functions can be ordered partially by
\begin{eqnarray*}
&{\cal I} \leq {\cal J} :\Longleftrightarrow \mbox{dom}({\cal I}) \subseteq \mbox{dom}({\cal J})\\
&\forall x\in \mbox{dom}({\cal I}):{\cal J}(x)={\cal I}(x)
\end{eqnarray*}

In addition to first order formulae, FIL provides formulae (so called
{\it ionic formulae}) of the form
$$
\ast (\{\phi_1,...,\phi_k\},\xi).
$$
Given a partial interpretation ${\cal I}$ that satisfies a set of
standard first order formulae $\Sigma$, the ionic formula above says
that $\xi$ is true in an extension of ${\cal I}$ as long as there is
no extension of ${\cal I}$ that assigns false to at least one $\phi_i$.

$\Phi=\{\phi_1,...,\phi_k\}$ is called {\it justification set} or
{\it justification context}. Stating the semantics of a {\it ionic formula}
more informally, we have that $\xi$ is true (by default) as long as
there is no evidence to the contrary from {\it justification context}
$\Phi$\footnote{In fact, \cite{abdallah} explains that an
interpretation function for $\Phi$ can assign 1 to $\Phi$ ($\Phi$ is
acceptable, written: $+\ast\Phi$), or 0 ($\Phi$ is inacceptable:
$-\ast \Phi$), or not assign 1 ($\Phi$ is not acceptable:
$\overline{+\ast}\Phi$), or not assign 0 ($\Phi$ is not inacceptable:
$\overline{-\ast}\Phi$). This is due to the fact that in a partial
logic for a relation symbol $R(x_1,...,x_n)$ one has two sets $R^{+}$
and $R^{-}$ to declare the semantics of $R(x_1,...,x_n)$:
$R(x_1,...,x_n)$ is true if and only if $(x_1,x...,x_n)\in R^{+}$ and
false if and only if $(x_1,x...,x_n)\in R^{-}$. $R^{+} \cup R^{-}$ do
not exploit the universe ${\cal U}$. So one can describe the four
``positions'' of $(x_1,...,x_n)$ relative to $R^{+}$ or $R^{-}$,
respectively.}.

\section{Partial Information and Dialog Plans}
\label{plans}
As outlined above, the main characteristic of dialogs is the fact that
information is given stepwise in the course of several utterances. For
the design of a domain-independent dialog manager it is therefore
important to develop an interpretation algorithm for utterances that
is able to interact with the user in order to collect the neccessary
information in any order.

{\it Ionic formulae} provide such a mechanism. Their {\it
justification contexts} allow for infering what information is still
missing while interpreting the current utterance. On the other hand, a
{\it justification context} can be seen as a set of default assumptions
to be accepted or rejected by the user later on.

From this point of view a {\it justification context} is the set of
dialog goals to be fulfilled by the dialog manager in order to compute
an answer to the user's question.

\section{Integrating FIL and DL}
\label{filtodl}
To combine the advantages of DL as a concept language for domain
modelling and FIL to describe partiality we integrate domain models
into the FIL based reasoning.

To do this, we have to characterize the part of a domain model that
can cause situations of partial information during a dialog. We state that
any role that has the concept {\sc User} as domain or range is a
source of ``missing information'' as these roles describe the user's
attitudes eventually to be clarified in several dialog
steps. Therefore, we define a set {\sc UserRel} as follows:
$$
\mbox{\sc UserRel}=\{R:\mbox{dom}(R)\subseteq\mbox{\sc User}\vee
\mbox{range}(R) \subseteq \mbox{\sc User}\}
$$
For a formal description of the connection of DL and FIL, we give
a translation mapping $\tau$ from DL to FIL which is sort of sensitive to
{\sc UserRel} defined above. The mapping, of course, resembles
Borgida's \cite{borgida} and the one by \cite{schmolzeisrael} and is
identical except of the names of roles:
$$
\tau(R)=\left\{
\begin{array}{ll}
\lambda\,x,y.\ast\!(\{R(x,y)\},\!R(x,y)),\, R\in\mbox{\sc UserRel}\\
\lambda\,x,y.R(x,y),\, \mbox{otherwise}
\end{array}
\right.
$$

This says that all role names of a given TBox
marked as partial are translated as ionic formulae.
All other symbols are translated to first order formulae as usual by
structural induction on the syntax of the DL. For example, given roles
$R$, $S_1$, and $S_2$ with $R=S_1\sqcup S_2$, the definition of $\tau$
is:
$$
\tau(R)=\lambda\,x,y.\tau(S_1)(x,y)\vee \tau(S_2)(x,y)
$$
whereas for role $R$ and concepts $C_1$ and $C_2$ with $C_1=\forall
R.C_2$ we have
$$
\tau(C_1)=
\lambda\,x.\forall y:\tau(R)(x,y)\longrightarrow\tau(C_2)(x)
$$
As it is shown by e.g.\ Borgida in \cite{borgida}, the mapping preserves
satisfiability of DL expressions that are translated into a FIL
formula, as FIL is an extension of First Order Predicate Logic and the
translation from FIL to FOPL only uses the FOPL ``sublanguage'' of
FIL. I.e.\ if a formula has some model in DL, then it has one in FIL, too.
On the other hand, if a FIL expression is satisfiable,
then it is in DL, too, if no ionic formula is contained (see
\cite{schmolzeisrael}).

A ionic formula $\ast(R(\alpha,\beta),R(\alpha,\beta))$ is the
translation of some role $R\in \mbox{\sc UserRel}$ whose {\it justification
context} $R(\alpha,\beta)$ can have one of the following states of
acceptance%\footnote{See the footnote above for a discussion of ``acceptance''.}:
\begin{itemize}
\item $+\ast R(\alpha,\beta)$ or $\overline{-\ast} R(\alpha,\beta)$:
I.e.\, for any model ${\cal M}$ it is impossible that ${\cal M}\false
R(\alpha,\beta) \leftrightarrow {\cal M}\models \neg R(\alpha,\beta)$. This observation implies that either ${\cal M}\models
R(\alpha,\beta)$ or $R(\alpha,\beta)$ is undefined. So
$\models \tau(\alpha ::R:\beta)=\ast(R(\alpha,\beta),R(\alpha,\beta))$
does not contradict $\models \alpha ::R:\beta$.
\item $-\ast R(\alpha,\beta)$ or $\overline{+\ast} R(\alpha,\beta)$:
in this case, analogously, either 
${\cal M}\false R(\alpha,\beta)$ or $R(\alpha,\beta)$ is undefined
implying that $\alpha ::\neg R:\beta$ is satisfiable.
\end{itemize}

In any case, $\tau$ preserves satisfiability depending on the state of
acceptance of {\it justification contexts}.

\section{Discussion of an Example}
\label{example}
In the train information application that is considered throughout this paper, an
appropriate terminology could include the following
concepts and roles:
\begin{itemize}
\item {\sc Train}, {\sc Depart}, {\sc Time}, {\sc Station}
\item $\mbox{\sc At}: \mbox{\sc Train}\times \mbox{\sc Time}$

$\mbox{\sc To}: \mbox{\sc Train}\times \mbox{\sc ArrStation}$

$\mbox{\sc From}: \mbox{\sc Train}\times \mbox{\sc DepStation}$

$\mbox{\sc DepartFrom}: \mbox{\sc User}\times \mbox{\sc Station}$
\end{itemize}
{\sc DepartFrom} is in {\sc UserRel} and therefore mapped as:
$$
\lambda\,u,s.\!\ast\!(\{\mbox{DepartFrom}(u,s)\},\!\mbox{DepartFrom}(u,s))
$$
Among many others we have the concepts
%{\footnotesize
\begin{eqnarray*}
\mbox{\sc DepStation}&=&\exists \mbox{\sc DepartFrom}^{-1}.\mbox{\sc
User}\\
&&\cap\,\mbox{\sc Station}\\
\mbox{\sc TrainFrom}&=&
\exists\mbox{\sc From}.\mbox{\sc DepStation}\\
&&\cap\,\mbox{\sc Train}\cap\mbox{\sc Depart}\\
\mbox{\sc TrainAtFrom}&=&
\exists\mbox{\sc At}.\mbox{\sc Time}\cap\mbox{\sc TrainFrom}\\
\mbox{\sc TrainAtFromTo}&=&
\exists\mbox{\sc To}.\mbox{\sc Station}\\
&&\cap\,\mbox{\sc TrainAtFrom}
\end{eqnarray*}%}
This terminology is translated to FIL follows (variables all-quantified):
\begin{eqnarray}
\nonumber \mbox{DepStation}(s)&\Longleftrightarrow &
\mbox{DepartFrom}^{-1}(s,u)\\
\label{depstat} && \wedge\,\mbox{User}(u)\wedge\,\mbox{Station}(s)\\
\nonumber \mbox{TrainFrom}(t)&\Longleftrightarrow&
\mbox{From}(t,s)\wedge \mbox{DepStation}(s)\\
&&\wedge\,\mbox{Train}(t)\wedge \mbox{Depart}(t)\\
\nonumber \mbox{TrainAtFrom}(t)&\Longleftrightarrow&
\mbox{At}(t,d)\wedge \mbox{Time}(d)\\
\label{atfrom} && \wedge\, \mbox{TrainFrom}(t)\\
\nonumber \mbox{TrainAtFromTo}(t)&\Longleftrightarrow&
\mbox{To}(t,s)\wedge \mbox{ArrStation}(s)\\
\label{atfromto} &&\wedge\,\mbox{TrainAtFrom}(t)
\end{eqnarray}
On the basis of the terminology above the utterance
\vskip.2\baselineskip
\centerline{{\it When does a train depart to Rome?}}
\vskip.2\baselineskip
\noindent has the discourse representation structure (DRS)
$$
\lambda x.\left[
\begin{array}{ll}
{\rm t}\,{\rm Rome}\\\hline
\mbox{Train}({\rm t})\\
\mbox{Depart}({\rm t})\\
\mbox{Time}(x)\\
\mbox{ArrStation}({\rm Rome})\\
\mbox{At}({\rm t},x)\\
\mbox{To}({\rm t},{\rm Rome})
\end{array}
\right]
$$

This structure is built up relying on DL reasoning: Rome is contained
in the (linguistic lexicon) as $\mbox{CityName}(\mbox{Rome})$. In this
example, the preposition {\it to} is mapped onto {\sc To} as defined
above. In order to combine {\it train}, {\it to}, and {\it Rome},
one has to check whether $\mbox{Rome}\in \exists \mbox{\sc
HasArrStation}.\mbox{\sc Station}$ which is evaluated by the data base
to be true for {\it Rome}. This results in
$\mbox{ArrStation}(\mbox{Rome})$. Generally, DRS are constructed by means of
instance checking that expands type unification.

To infer $\mbox{TrainFrom}({\rm t})$ on the basis of the information
available from the utterance, it is necessary to determine the truth
value of $\phi(s)=\lambda s.\mbox{DepartFrom}^{-1}(s,{\rm u})$ (see
eq.\ \ref{depstat}; variable $u$ is bound to constant $\mbox{u}\in
\mbox{\sc User}$). A first order approach would answer false, because its
interpretation function is total and there is no information in the
DRS above in order to substitute a station name for $s$ (making $\phi$
true). But in FIL we have (see above):
\begin{eqnarray*}
\mbox{DepartFrom}^{-1}(s,u)&\Longleftrightarrow&\ast\!\,(\{\mbox{DepartFrom}(u,s)\},\\
&&\!\mbox{DepartFrom}(u,s))
\end{eqnarray*}

Based on this rule, we can conclude immediately that $\phi(s)$ be true
iff $\xi(s)=\lambda
s.\mbox{DepartFrom}({\rm u},s)$ is true unless there is information to
the contrary (see Sect.\ \ref{fil}).

As $\xi$'s {\it justification context} still contains an unbound
variable, the dialog manager interprets it as a question to be posed
to the user (no default assumptions can be made). Therefore, the
discourse plan will be updated and the dialog manager will react with
\vskip.15\baselineskip
\centerline{{\it Where do you depart to from?}}
\vskip.15\baselineskip
because $s\in \mbox{\sc Station}$. The answer
\vskip.15\baselineskip
\centerline{{\it From Milan.}}
\vskip.15\baselineskip
\noindent adds $\mbox{DepartFrom}(\mbox{u},\mbox{Milan})$ to the {\it
shared knowledge} so that the dialog manager can bind $s$ to Milan.
After that, we can infer by eq.\ \ref{atfrom} and eq.\ \ref{atfromto}
$\mbox{TrainAtFromTo}(\mbox{t})$ substituting $x$ by $d$ and $d$ by
all constants $c$ for whom $\mbox{At}(\mbox{t},\mbox{c}) \wedge
\mbox{Time}(\mbox{c})$ is true.

The dialog plan is based on a speech act model\footnote{For
information dialogs we assume as minimal the set \{{\sf inform}, {\sf
query}, {\sf suggest}, {\sf accept}, {\sf reject}\}. This set can be
extended to fit the needs of a certain application. See
\cite{konvens98} for details.}. Speech acts are
determined by reasoning on the user's attitudes, grammatical
information from, and coherence of utterances.

\section{From Notions to Vocabulary}
To make a dialog system understand the user it has to know how
expressions in natural language are connected to the abstract notions
of the domain modell.

Natural languages normally offer the possibility to express the same
notion by different synonyms. For example, one can say:
{\it When does the train leave to Rome?}
or, alternatively {\it When does the train depart to Rome?}
In both cases, the notion of {\it train departure} is expressed.

When we inspected the EVAR\footnote{EVAR is a publicly accessible
information system on German Railway InterCity connections
(\cite{EVAR}). The existing corpus of more than 1100 annotated dialogs
contains samples of ``real world'' data with ``naive'' users.} corpus of train
information dialogs, we could see that synonyms occur frequently, even in
relatively simple domains as train information. Our hypothesis is that
in much larger more complicated domains that allow for a greater
variability of expressions and vocabulary there are even more synonyms
for a certain abstract notion.
 
To handle this phenomenon, we have to extend our domain description by
adding formulae like 
$$
\mbox{Syn}_1^C(x) \vee \cdots \vee \mbox{Syn}_n^C(x)
\Longrightarrow C(x)
$$
for any concept $C$ and and its synonymic expressions
$\mbox{Syn}_i^C(x)$ and, equally,
$$
\mbox{Syn}_1^R(x,y) \vee \cdots \vee
\mbox{Syn}_n^R(x,y) \Longrightarrow R(x,y)
$$
for any role $R$ and its synonyms $\mbox{Syn}_i^R(x,y)$.

For the two questions above we would introduce
$$
\mbox{leave}(x)\vee \mbox{depart}(x)\Longrightarrow \mbox{\sc Depart}(x)
$$
``mapping'' the verbs {\it leave} and {\it depart} to {\sc Depart}.
 
\section{Pragmatics of Concepts and Roles}

Except of constructing a linkage between a domain model in terms of DL
and a language model for the given domain, a configurable dialog manager
must define an interface between its logical domain model and an
arbitrary (mostly non-logical) problem solving component for the domain.

The reasoning mechanisms of the dialog manager and the
problem solver, respectively, can be linked as follows: The problem
solver evaluates relations between (i.e.\ roles of) discourse
referents that are instantiations of concepts according to the
previous utterances.

In the example outlined above, t can be asserted $\mbox{\sc TrainAtFromTo}$
only if the query
$$
\mbox{At}(\mbox{t},x)\wedge \mbox{From}(\mbox{t},\mbox{Milan})\wedge
\mbox{To}(\mbox{t},\mbox{Rome})
$$
can be evaluated successfully (e.g.\ as a query to a database) and
returns a list of connections from Milan to Rome. They can serve as
data for continuing the dialog.

In the way outlined we can define an interface between dialog management and
the pragmatics of the application which is by nature independent of a
specific domain and therefore allows for abstraction of the dialog
manager from its underlying domain-dependent problem solving
component, linguistics, and discourse structure.

\section{Configuring a Dialog Manager}
\label{configure}
For the practical purpose of adapting a dialog manager for a specific
task it is of great importance to observe that dialogs consist of a
domain-independent and domain-dependent utterances.

Domain independence is related to establishing mutual understanding,
dialog segmentation, and reference resolution. It is very important to
consider these phenomena as they can be expressed in natural language:
e.g.\ ``Could you repeat that?'', ``Pardon?'', ``Next I want to ask
you ...''. Not to take such
expressions into account would result in poor understanding
capabilities of the dialog manager. Intentions, speech acts, and
obligations can be expressed explicitly as well\footnote{By modifying
this dialog model, one can influence the planning of the dialog
manager (c.f.\ footnote in Sect.\ \ref{example}).}.

A model of these domain independent notions forms the basic
capabilities of the dialog manager to engage in natural language
conversation. To configure it for a certain application, one has to
expand the model describing application-specific (i.e.\
domain-dependent) notions by
\begin{enumerate}
\setlength{\itemsep}{0pt}
\setlength{\parsep}{0pt}
\item Adapting, extending, and specializing the given DL model so that
it defines all notions of importance for the application. DL systems
support the phase of designing a domain model by means of testing
the satisfiability of terminologies. This is a major practical advantage
for ensuring the robustness of the dialog manager compared to other
approaches to domain modelling.

\item Defining the interface between the problem solving component and
the dialog manager. I.e.\ defining what concept and role symbols will be
evaluated by what functions of the problem solver.
\end{enumerate}
\section{Conclusions}
We have established a connection between DL and FIL in terms of
satisfiability via a mapping between formulae of each
language. The correctness of inferences is assured by FIL (see
\cite{abdallah}). This allows to conclude the correctness of reasoning
with DRS as they can be mapped onto FIL formulae (see \cite{reyle} and
\ref{drtanddl}). We consider the combination of DL's knowledge representation
facilities and FIL's inference mechanism for partial knowledge a
well-founded basis for utterance interpretation and the description of
mixed initiative dialogs in order to implement the ``core engine'' of
an adaptive, configurable, and domain independent dialog manager. In
this way we can separate linguistics and discourse theory
from the knowledge engineering task to describe the application
domain. This task can be performed by an application expert even without
deep knowledge of dialog managers.

\section{Future Research}
 Evidence from a number of experiments  shows that humans perform
domain reasoning while incrementally matching the speaker's utterances
with their own expectations (see e.g.\
\cite{Poesio,Traum}). Following this line of research, we are studying
the use of domain descriptions and DL reasoning services to model how dialog
participants establish mutual beliefs on the basis of utterances.

Secondly, cooperating industrial partners we want to generate
domain models for different domains. For the acquisition of ``synonym
knowledge'' we will apply learning techniques from corpora of example
dialogs collected by our partners. General work on DL learning (see
e.g.\ \cite{frazier}) as well as work on the combination of DL and
linguistic processing (as done in \cite{schnaittinger}) will have to
be considered. 

Finally, we are working on user-friendly tools to define interfaces
between domain models and its related problem solving components.

\section{Acknowledgements}
We want to thank two anonymous reviewers for their comments that
helped to improve this paper.


\begin{thebibliography}{10}
\footnotesize \parskip=0pt \itemsep=0pt
\bibitem{abdallah}
{\bf Abdallah, N.:} {\it The Logic of Partial Information}, Springer,
New York 1995
\bibitem{agarwal}
{\bf Agarwal, R.:} {\it Towards a PURE Spoken Dialogue System for
  Information Access}, in: {\it Proceedings of the ACL/EACL Workshop on Interactive Spoken Dialog Systems:
       Bringing Speech and NLP Together in Real Applications}, Madrid, Spain, pp. 90--97, 1997
\bibitem{borgida}
{\bf Borgida, A.:} {\it On the Relative Expressiveness of
Description Logics and Predicate Logics}, Artificial Intelligence,
volume 82 (1996), pp.\ 353--367
\bibitem{Eco1993}
{\bf Eco, U.:} {\it La ricerca della lingua perfetta nella cultura
europea}, Laterza, Roma, Bari 1993
\bibitem{EVAR}
{\bf Eckert, W., Niemann, H.:} {\it Semantic Analysis in a
  Robust Spoken Dialog System}, in: {\it Proc.\
       Int.\ Conf.\ on Spoken Language Processing}, 
       Yokohama, 1994, pp.\ 107--110
\bibitem{frazier}
{\bf Frazier, M.\ and Pitt, L.:} {\it CLASSIC learning},
Machine Learning, 25 (1996), pp.\ 151--194
\bibitem{Hjelmslev1943}
{\bf Hjelmslev, L.:}, {\it Omkring sprogteoriens grundlaeggelse},
Munksgaard, Kopenhagen 1943; engl.\ {\it Prolegomena to a Theory of
Language}, Univ.\ of Wisconsin Press, 1961
\bibitem{shriberg}
{\bf Jurafsky, D., et al.:} {\it Automatic Detection of Discourse Structure
for Speech Recognition and Understanding.}, in:  {\it Proc. 1997 IEEE
Workshop on Speech Recognition and Understanding}, Santa Barbara, 1997
\bibitem{reyle}
{\bf Kamp, H.\ and Reyle, U.:} {\it From Discourse to Logic},
Kluwer, Dordrecht 1993
\bibitem{baekgaard} {\bf Larsen, L.\ B.:} {\it Development and
    Evaluation of a Spoken Dialogue for a Telephone Based Transaction
    System}, in: {\it Proc.\ EUROSPEECH 95}, pp.\ 1973--1976
%\bibitem{BL}
%{\bf Ludwig, Bernd:} {\it Partielle Logik in Semantik und Diskurs},
%Diplomarbeit, Universit\"at Erlangen-N\"urnberg 1997
\bibitem{konvens98}
{\bf Ludwig, B., G\"orz, G., Niemann, H.:} {\it User Models, Dialog Structure, and Intentions in Spoken Dialog}, submitted to KONVENS 98
\bibitem{moeller}
{\bf Moeller, J.-U.:} {\it DIA-MOLE: An Unsupervised Learning Approach to Adaptive Dialogue Models for Spoken
       Dialogue Systems}, in: {\it Proc. EUROSPEECH 97}, pp.\ 2271-2274
\bibitem{novick}
{\bf Ward, K.\ and Novick, D.\ G.:} {\it On the Need for a Theory of Integration of Knowledge Sources for
Spoken Language Understanding}, in: {\it Proceedings of the AAAI-94
Workshop on the Integration of Natural Language and Speech Processing}, Seattle, 1994, pp.\ 23--30
\bibitem{Poesio}
{\bf Poesio, M.:} {\it Discourse Interpretation and the Scope of
  Operators}, Ph.D.\ Thesis, Computer Science Dept., University of
Rochester 1994
\bibitem{Russell1940}
{\bf Russell, B.:} {\it The Object Language}, in: Russell, B.:
{\it An Enquiry into Meaning and Truth}, Allen and Unwin, London 1950
\bibitem{schmolzeisrael}
{\bf Schmolze, J.G.\ and Israel, D.:} {\it KL-ONE: Semantics and
Classification}, in {\it Research in Knowledge Representation for NL
Understanding}, Tech Report 5421, BBN Laboratories 1983
\bibitem{schnaittinger}
{\bf Schnaittinger, K.\ and Hahn, U.:} {\it Constraining the
  Acquisition of Concepts by the Quality of Heterogeneous Evidence},
in: {\it Proceedings of the 21$^{\rm st}$ Annual German Conference on
  Artificial Intelligence} (LNAI 1303), Springer, Berlin 1997, pp.\ 255--266
\bibitem{Traum}
{\bf Traum, D.:} {\it A Computational Theory of Grounding in
  Natural Language Conversation}, Ph.D.\ Thesis, Computer Science
Dept., University of Rochester 1994
\end{thebibliography}
\end{document}